\documentclass[11pt]{article}

\usepackage[preprint]{acl}

\usepackage{times}
\usepackage{latexsym}

\usepackage[T1]{fontenc}

\usepackage[utf8]{inputenc}

\usepackage{microtype}

\usepackage{inconsolata}

\usepackage{graphicx}

\usepackage{tcolorbox}
\usepackage{verbatim}
\usepackage{booktabs}
\usepackage{threeparttable}
\usepackage{makecell}
\usepackage{pifont}    
\usepackage{placeins} 

\usepackage[table]{xcolor}
\definecolor{lightblue}{RGB}{235,242,249}
\definecolor{lightred}{RGB}{250,230,235}
\definecolor{jpblue}{RGB}{33, 99, 154}
\definecolor{jpred}{RGB}{188, 0, 45}

\usepackage{xspace}
\usepackage{gradient-text}
\usepackage{CJKutf8}
\usepackage{amsmath}

\newcommand{\ebisu}{\gradientRGB{Ebisu}{33, 99, 154}{188, 0, 45}\xspace}

\title{\textsc{\ebisu}: Benchmarking Large Language Models in Japanese Finance}

\author{
\begin{tabular}{c}
\textbf{Xueqing Peng\textsuperscript{2}},
\textbf{Ruoyu Xiang\textsuperscript{3}},
\textbf{Fan Zhang\textsuperscript{4}},
\textbf{Mingzi Song\textsuperscript{5}}, \\
\textbf{Mingyang Jiang\textsuperscript{2}},
\textbf{Yan Wang\textsuperscript{2}},
\textbf{Lingfei Qian\textsuperscript{2}},
\textbf{Taiki Hara\textsuperscript{1}},
\textbf{Yuqing Guo\textsuperscript{2}}, \\
\textbf{Jimin Huang\textsuperscript{1,2}},
\textbf{Junichi Tsujii\textsuperscript{6}},
\textbf{Sophia Ananiadou\textsuperscript{1,7}}
\end{tabular}
\\[10pt]
\\
\begin{tabular}{c}
\textsuperscript{1}\textit{University of Manchester},
\textsuperscript{2}\textit{The Fin AI},
\textsuperscript{3}\textit{New York University}, \\
\textsuperscript{4}\textit{The University of Tokyo},
\textsuperscript{5}\textit{Meiji Gakuin University}, \\
\textsuperscript{6}\textit{National Institute of Advanced Industrial Science and Technology (AIST)}, \\
\textsuperscript{7}\textit{The National Centre for Text Mining}
\end{tabular}
\\[10pt] 
\\
\small{\textbf{Correspondence:} \texttt{jimin.huang@postgrad.manchester.ac.uk}}
}

\begin{document}
\maketitle
\begin{abstract}
Japanese finance combines agglutinative, head-final linguistic structure, mixed writing systems, and high-context communication norms that rely on indirect expression and implicit commitment, posing a substantial challenge for LLMs. We introduce \textsc{\ebisu}, a benchmark for native Japanese financial language understanding, comprising two linguistically and culturally grounded, expert-annotated tasks: \textit{JF-ICR}, which evaluates implicit commitment and refusal recognition in investor-facing Q\&A, and \textit{JF-TE}, which assesses hierarchical extraction and ranking of nested financial terminology from professional disclosures. We evaluate a diverse set of open-source and proprietary LLMs spanning general-purpose, Japanese-adapted, and financial models. Results show that even state-of-the-art systems struggle on both tasks. While increased model scale yields limited improvements, language- and domain-specific adaptation does not reliably improve performance, leaving substantial gaps unresolved. \textsc{\ebisu} provides a focused benchmark for advancing linguistically and culturally grounded financial NLP. All datasets and evaluation scripts are publicly released.
\footnote{\tiny
Data:\hspace*{0.4em}\url{https://huggingface.co/datasets/TheFinAI/JF-ICR}\\
\hspace*{5.3em}\url{https://huggingface.co/datasets/TheFinAI/JF-TE}\\
\hspace*{2.7em}Code:\hspace*{0.3em}\url{https://github.com/The-FinAI/Ebisu}
}
\end{abstract}

\begin{figure}[ht]
  \includegraphics[width=1.0\linewidth]{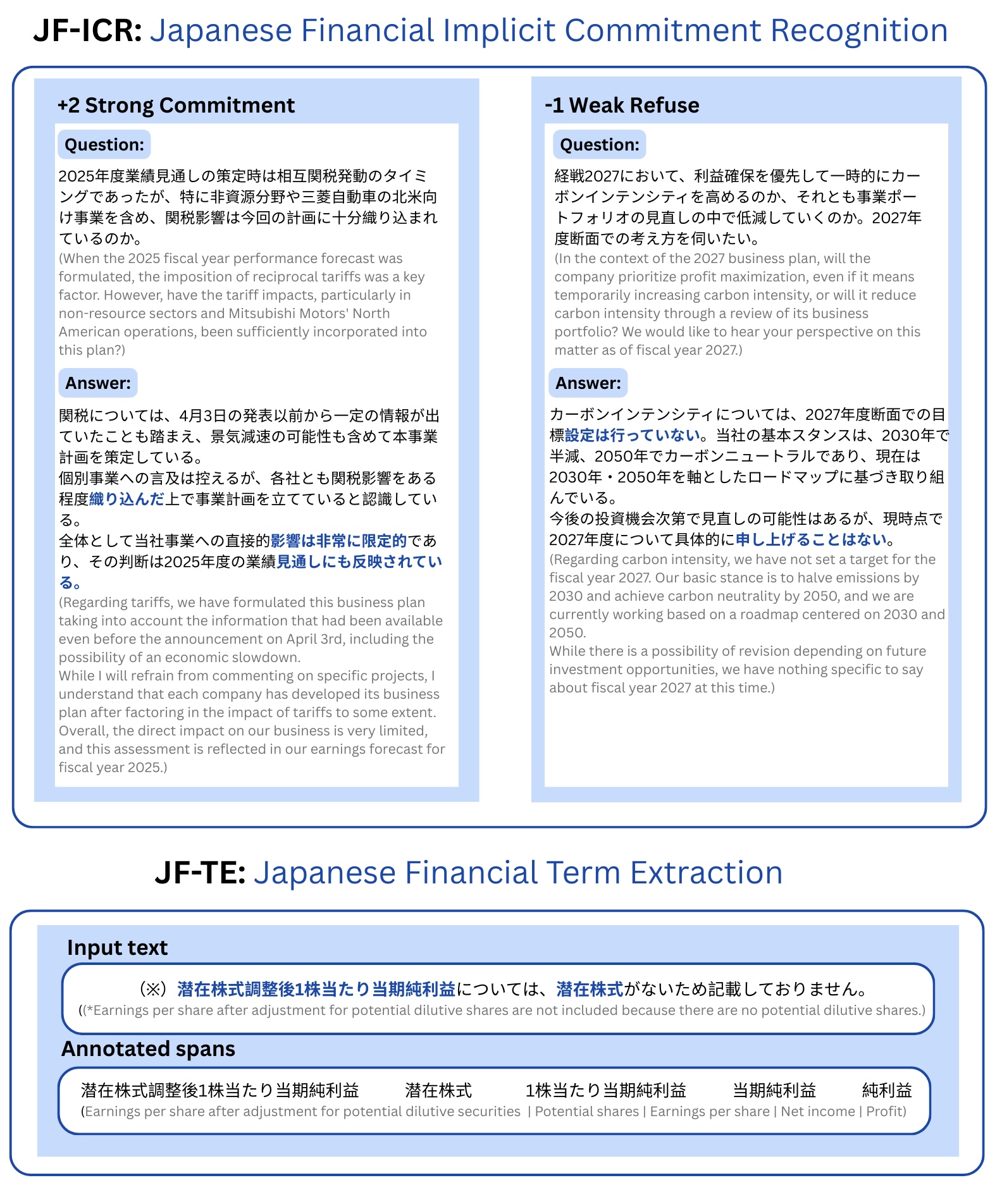}
  \caption {Representation examples of \textit{JF-ICR} and \textit{JF-TE}.}
  \label{fig_datasets}
\end{figure}

\DeclareRobustCommand{\cmark}{\textcolor{green!50!black}{\ding{51}}}
\DeclareRobustCommand{\xmark}{\textcolor{red!70!black}{\ding{55}}}
\DeclareRobustCommand{\na}{\textcolor{gray}{--}}

\begin{table}[!htbp]
\centering
\scriptsize
\renewcommand{\arraystretch}{1.05}
\resizebox{\columnwidth}{!}{
\begin{tabular}{lccccc}
\toprule
\textbf{Benchmark}\strut &
\textbf{JA}\strut &
\textbf{Fin}\strut &
\textbf{Ling}\strut &
\textbf{Prag}\strut &
\textbf{FC}\strut \\
\midrule
JGLUE~\citep{kurihara-etal-2022-jglue} & \cmark & \xmark & \xmark & \xmark & \xmark \\
JBLiMP~\citep{someya-oseki-2023-jblimp} & \cmark & \xmark & \cmark & \xmark & \xmark \\
FinBen~\citep{NEURIPS2024_adb1d9fa} & \xmark & \cmark & \xmark & \xmark & \xmark \\
FinQA~\citep{chen2022finqadatasetnumericalreasoning} & \xmark & \cmark & \xmark & \xmark & \xmark \\
MultiFinBen~\citep{peng2025multifinbenbenchmarkinglargelanguage} & \cmark & \cmark & \xmark & \xmark & \cmark \\
Japanese-LM-Fin-Harness~\citep{hirano2024japanese_lm_fin_harness} & \cmark & \cmark & \xmark & \xmark & \xmark \\
EDINET-Bench~\citep{sugiura2025edinet_bench} & \cmark & \cmark & \xmark & \xmark & \xmark \\
\rowcolor{lightblue}
\textbf{\textsc{\ebisu} (ours)} & \cmark & \cmark & \cmark & \cmark & \cmark \\
\bottomrule
\end{tabular}
}
\caption{Comparison with existing benchmarks. JA: Japanese language coverage; Fin: financial domain coverage; Ling: linguistic phenomena; Prag: pragmatic understanding; FC: financial--cultural knowledge.}
\label{tab:bench_positioning}
\end{table}

\section{Introduction}

Even state-of-the-art large language models (LLMs) can get lost in Japanese finance. This is linguistically grounded because Japanese is agglutinative and head final~\citep{mccawley1975structure}, so modality and negation are often realized sentence finally~\citep{hasegawa2018cambridge,narrog2007modality} and refusals are frequently conveyed indirectly through pragmatic cues~\citep{ide1989formal}. Interpretation is further complicated by Japanese financial term formation, where nested nominal compounds and mixed script variants (kanji, hiragana, and katakana~\citep{bond2016introduction,joyce2017constructing,tamaoka2014japanese}) make semantic scope depend on term boundaries and reference. Since Japanese corporate disclosures and investor communications are already consumed and processed at global scale, evaluating this capability matters for cross border finance rather than only domestic use, especially given that foreign investors held 32.4\% of listed share market value in Japan in FY2024~\citep{bloomberg2025_japan_foreign_share}.


However, existing multilingual and financial LLM evaluations are optimized for cross-language comparability and scalable supervision~\citep{peng2025multifinbenbenchmarkinglargelanguage, NEURIPS2024_adb1d9fa}, which biases them toward short contexts, explicit targets, and label schemes that remain stable across languages and annotators. Multilingual benchmarks therefore concentrate on QA, NLI, and multiple-choice style formats~\citep{ qian2025agents,zhang2024dolares,peng2025plutus}, while finance benchmarks are typically instantiated as canonical task templates with clear input–output supervision~\citep{wang2025fintagging, wang2025finauditing,qian2025fino1}, making both less suited to phenomena that are hard to operationalize at scale, such as discourse-level intent conveyed indirectly and stance resolved by sentence-final morphology~\citep{nasukawa2025empathetic}. Japanese finance resources~\citep{hirano2024construction} often inherit the same constraints because high-quality native financial text is difficult to curate and license and pragmatic intent labels are costly to annotate reliably, so available datasets tend to emphasize sentiment~\citep{nakatsuka2025extracting} or exam-style QA~\citep{sugiura2025edinet_bench}.

In this work, we introduce \textsc{\ebisu}, a benchmark for native Japanese financial language understanding that targets two coupled requirements in real corporate communication, stance inference about commitment and refusal and boundary-sensitive grounding of finance terminology in mixed-script text (Table~\ref{tab:ebisu-benchmark}). \textit{JF-ICR} (Japanese Financial Implicit Commitment Recognition) consists of 94 investor-facing Q\&A pairs from 4 companies spanning 3 years, annotated into 5 implicit stance categories that distinguish agreement, hedging, and refusal under high-context discourse. \textit{JF-TE} (Japanese Financial Term Extraction) consists of 202 note-level instances extracted from 10 professional disclosures with 2,412 expert-curated term mentions covering 777 unique finance terms after normalization, where nested compounds and script-variant loanwords make term boundaries and semantic scope ambiguous. Both datasets are labeled by finance-trained native-level Japanese annotators following a detailed guideline and each item is independently double-annotated, disagreements are adjudicated by a senior expert, and we report inter-annotator agreement (Table~\ref{tab:annotation-agreement}) together with audit-based quality checks on stance consistency and term-boundary validity. We evaluate \textit{JF-ICR} using Accuracy as the primary metric and \textit{JF-TE} as a ranked retrieval task using F1 and HitRate@K (K=[1/5/10]) under exact matching, enabling diagnostic evaluation of whether models recover what is being discussed and what is being committed to in Japanese financial discourse.

We evaluate 22 LLMs on \textsc{\ebisu}, including both open-weight and proprietary systems, and find that overall performance remains low across models, highlighting the challenging nature of the benchmark. We conduct within-family, size-matched comparisons to analyze the effects of model scale, Japanese language adaptation, and financial domain adaptation. Increasing model scale consistently improves performance on both \textit{JF-ICR} (+0.33 Acc) and \textit{JF-TE} (+0.38 F1), but these gains remain limited. In contrast, language- and domain-specific adaptation does not reliably yield improvements: Japanese-adapted models do not consistently outperform their English counterparts at matched scale, and continued financial pretraining can even degrade performance, particularly for financial term extraction (-0.12 F1). Across both tasks, errors concentrate on linguistically and pragmatically defined phenomena rather than missing domain knowledge. On \textit{JF-ICR}, models struggle with cases where stance is resolved through clause-final auxiliaries or indirect refusal strategies, frequently confusing refusal with neutral or hedged responses. On \textit{JF-TE}, ranking quality deteriorates with increasing compound depth and mixed-script variability, and the gap between HitRate@1 and HitRate@10 indicates difficulties in term boundary resolution and variant handling rather than lack of financial vocabulary. Together, these findings suggest that current scaling and adaptation strategies improve average accuracy without reliably capturing the morphosyntactic and pragmatic mechanisms central to Japanese financial communication.

Our contributions are fourfold:
\begin{itemize}
    \item \textbf{\textsc{\ebisu} benchmark.} We introduce \textsc{\ebisu}, a benchmark for native Japanese financial language understanding that targets pragmatic stance inference and finance-term grounding under Japanese-specific linguistic and discourse conditions.
    
    \item \textbf{New tasks and datasets grounded in Japanese finance.} We define two new expert-annotated tasks and datasets that deeply reflect Japanese financial practices: \textit{JF-ICR} for implicit commitment recognition in investor-facing Q\&A and \textit{JF-TE} for extraction and ranking of finance-relevant terminology in professional disclosures. 
    
    \item \textbf{Comprehensive evaluation and diagnostic analysis.} We conduct a broad evaluation over 22 LLMs and provide diagnostic analyses that isolate the effects of scaling and adaptation, showing that improvements are limited and that strong performance is not guaranteed by proprietary training alone.
    
    \item \textbf{Public resources.} We release \textsc{Ebisu} and accompanying annotation guidelines and evaluation code to facilitate language- and culture-aware financial NLP research.
\end{itemize}

\section{Related Work}
\subsection{Japanese Financial LLMs}
The development of Japanese LLMs has transitioned from general-purpose adaptation to deep financial specialization. Early efforts like the Swallow series \citep{fujii2024swallow} established a successful pipeline for enhancing Japanese syntax and reasoning through continual pre-training and vocabulary expansion. Financial domain alignment was subsequently advanced by works like \cite{tanabe2024jafin, hirano2024domain}, which provides high-quality instructions sourced from authoritative entities like the Bank of Japan and the Financial Services Agency to ensure model responses adhere to professional regulatory standards. Industry innovations include SoftBank's Sarashina\footnote{https://huggingface.co/sbintuitions/sarashina2.1-1b}, which targets "sovereign AI" with 460 billion parameters for legal language understanding, and Stockmark-LLM\footnote{https://huggingface.co/stockmark/stockmark-100b}, which suppresses hallucinations in business contexts through training on extensive news and patent corpora.

\subsection{Japanese Financial Benchmarks}

Evaluation frameworks for Japanese financial LLMs have evolved from knowledge-based tests to expert-level reasoning tasks. The Japanese-LM-Fin-Harness~\citep{hirano2024japanese_lm_fin_harness} first provided an initial substrate for benchmarking by incorporating securities analyst exams, CPA audit theory, and binary sentiment analysis. Recently, to capture expert-level reasoning, EDINET-Bench \citep{sugiura2025edinet_bench} introduced expert tasks such as accounting fraud detection and earnings forecasting grounded in a decade of official filings. Specialized assessments like SIG-FIN UFO-2024 \cite{kimura2024understanding} focus on table retrieval and QA within securities reports, while pfgen-bench \cite{jiantailangpfgen} measures the quality of financial text generation. Other studies on small-cap stocks \cite{suzuki2025sentiment} and earnings briefing dialogues \cite{nakatsuka2025extracting} further highlight that specialized LLMs can identify predictive signals and nuanced investor stances that traditional methods overlook. 
Existing evaluations span general Japanese linguistic benchmarks \citep{kurihara-etal-2022-jglue,someya-oseki-2023-jblimp} and global financial benchmarks \citep{NEURIPS2024_adb1d9fa,chen2022finqadatasetnumericalreasoning}, but the former lacks finance coverage and the latter is largely English-centric. Cross-lingual efforts \citep{peng2025multifinbenbenchmarkinglargelanguage} partly bridge Japanese finance, yet often remain too coarse for fine-grained linguistic analysis.

\section{\textsc{\ebisu} Benchmark}

In this section, we introduce \textsc{\ebisu}, a benchmark designed to evaluate LLMs on native Japanese financial language understanding. \textsc{\ebisu} consists of two complementary tasks targeting (1) Japanese Financial Implicit Commitment Recognition (\textit{JF-ICR}) in high-context Japanese financial communication and (2) Japanese Financial Term Extraction (\textit{JF-TE}) from professional financial documents (Table~\ref{tab:ebisu-benchmark}). 

\begin{table*}[!htbp]
\renewcommand{\arraystretch}{1}
\centering
\small
\begin{threeparttable}
\resizebox{\textwidth}{!}{
\begin{tabular}{@{}lccccccc@{}}
\toprule
\textbf{Task / Dataset} & \textbf{Source} & \textbf{Test Size} & \textbf{Metrics} & \textbf{License} & \textbf{Tested Capabilities} \\ 
\midrule
\textit{JF-ICR}
& Company Q\&A\tnote{1} 
& 94
& Accuracy 
& Public 
& Implicit intent understanding in Japanese finance \\

\textit{JF-TE}
& Annual Securities Reports\tnote{2} 
& 202
& F1 and HitRate@K\tnote{3} 
& Public 
& Financial terminology extraction and ranking \\

\bottomrule
\end{tabular}}
\begin{tablenotes}
\tiny
\item[1] \url{https://www.mufg.jp/ir/presentation/2024/index.html}; \url{https://www.smfg.co.jp/investor/financial/presentation.html}; \url{https://www.itochu.co.jp/ja/ir/} \\
\url{financial_statements/2025/index.html}; \url{https://www.mufg.jp/ir/stock/meeting/2023/}; \url{https://www.mitsubishicorp.com/jp/ja/ir/}
\item[2] \url{https://disclosure2.edinet-fsa.go.jp/}
\item[3] We report HitRate@1, HitRate@5, and HitRate@10 to evaluate the accuracy of financial terms at different ranking depths.
\end{tablenotes}
\caption{Overview of the \textsc{\ebisu} benchmark. For each task, we report the data source, data size, evaluation metrics, license information, and the targeted capabilities.}
\label{tab:ebisu-benchmark}
\end{threeparttable}
\end{table*}

\subsection{\textit{JF-ICR}: Japanese Financial Implicit Commitment Recognition}

Japanese communication is characterized by indirectness, politeness, and implicit expression of intent, making agreement or refusal difficult to infer. This difficulty is compounded by the agglutinative nature of Japanese, where critical semantic and pragmatic information is often encoded at the sentence-final position, in sharp contrast to English or Chinese, leading to frequent errors in polarity, commitment, and risk interpretation. In Japanese financial discourse, intent inference becomes even more challenging due to the prevalent use of vague and abstract expressions, limited explicit negative assertions, hedged and hypothetical writing styles, layered negation, and katakana-based loanwords whose meanings may diverge from their source languages, often framing potential risks in a neutral or favorable light. 

As most large language models are predominantly trained on English-centric corpora and U.S.-style financial disclosures, they may struggle to accurately infer the true intent behind Japanese financial communication. The \textit{JF-ICR} task is designed to evaluate this capability by assessing whether models can correctly identify implicit agreement and refusal in such high-context settings.

\paragraph{Task Definition.}

The \textit{JF-ICR} task is formulated as an intent classification problem over Japanese financial communication. 
Given a single-turn financial question–answer pair \( (q, a) \), where \( q \) denotes a financial query and \( a \) denotes the corresponding company response in Japanese, the model is required to infer the underlying intent expressed in the response \( a \).
The output is a five-level intent label \( y \in \{-2, -1, 0, +1, +2\} \), corresponding to \textit{Strong Commitment} (\(+2\)), \textit{Weak or Qualified Commitment} (\(+1\)), \textit{Neutral or Hedged Intent} (\(0\)), \textit{Weak Refusal} (\(-1\)), and \textit{Strong Refusal} (\(-2\)). This formulation focuses on intent inference from the response \( a \), rather than sentiment classification or surface-level polarity detection.

\paragraph{Data Source.}

We construct \textit{JF-ICR} from real-world, publicly available Japanese corporate disclosures in which companies respond to financial inquiries in high-context settings. Specifically, we collect question--answer transcripts from
(i) earnings call / investor briefing Q\&A\footnote{
\tiny
\begin{tabular}[t]{@{}l@{}}
\url{https://www.mufg.jp/ir/presentation/2024/index.html}\\
\url{https://www.smfg.co.jp/investor/financial/presentation.html}\\
\url{https://www.itochu.co.jp/ja/ir/financial_statements/2025/index.html}
\end{tabular}},
(ii) shareholder meeting Q\&A\footnote{
\tiny
\begin{tabular}[t]{@{}l@{}}
\url{https://www.mufg.jp/ir/stock/meeting/2023/}
\end{tabular}},
and (iii) financial results briefings (\textit{kessan setsumeikai}) Q\&A\footnote{
\tiny
\begin{tabular}[t]{@{}l@{}}
\url{https://www.mitsubishicorp.com/jp/ja/ir}
\end{tabular}}.
In total, our corpus comprises 8 source documents from 4 companies spanning 2023–2026, with an average of 10 Q\&A exchanges per document.

To ensure unambiguous intent annotation, we perform careful manual curation. We retain only single-turn question--answer pairs that focus on a single topic, and exclude multi-part questions, intertwined follow-ups, or exchanges that require broader conversational context. This filtering process reduces annotation ambiguity and improves data quality. After curation, we obtain 94 single-turn Japanese financial Q\&A instances for expert annotation.

\paragraph{Expert Annotation.}

We developed a rigorous annotation guideline (Appendix~\ref{sec:app_icr_ag}) in collaboration with Japanese financial experts. Each instance was annotated using the five-point agreement/refusal scale \(\{+2, +1, 0, -1, -2\}\), with detailed criteria specified for each label to ensure consistency. The guidelines were iteratively refined through pre-annotation rounds, with particular attention to incorporating rules for ambiguous cases. Only responses expressing \textit{strong, clear, explicit, and definitive} agreement or refusal were assigned the \(\pm 2\) labels.
All annotations were performed by 2 native-level Japanese financial experts with extensive industry experience (Appendix~\ref{sec:app_demography}). The annotation process was conducted using the Label Studio platform (Figure~\ref{fig_labelstudio_icr}, Appendix~\ref{sec:app_labelstudio}), ensuring a streamlined and reproducible workflow.

\paragraph{Quality Validation.}

To assess the reliability of the \textit{JF-ICR} annotations, we measure inter-annotator agreement using standard metrics for multi-class classification tasks, including Macro-F1~\citep{sokolova2009systematic}, Cohen’s $\kappa$~\citep{cohen1960kappa}, and Krippendorff’s $\alpha$~\citep{krippendorff2011alpha} (Appendix~\ref{sec:app_validation}). Macro-F1 captures balanced agreement across the five intent categories, while Cohen’s Kappa and Krippendorff’s Alpha adjust for chance agreement and label distribution skew.
The agreement results indicate a high level of consistency between annotators (Table~\ref{tab:annotation-agreement}), demonstrating that the annotation guidelines are clearly defined and consistently applied. 

\begin{table}[t]
\centering
\scriptsize
\begin{tabular*}{\linewidth}{l@{\extracolsep{\fill}}ccc}
\toprule
\textbf{Dataset} & \textbf{Macro-F1} & \textbf{Cohen’s $\kappa$} & \textbf{Krippendorff’s $\alpha$} \\
\midrule
\textit{JF-ICR} & 0.9215 & 0.8769 & 0.8768 \\
\textit{JF-TE}  & 0.8230 & 0.7786 & 0.7832 \\
\bottomrule
\end{tabular*}
\caption{Inter-annotator agreement results for \textit{JF-ICR} and \textit{JF-TE}, evaluated using Macro-F1, Cohen’s $\kappa$, and Krippendorff’s $\alpha$.}
\label{tab:annotation-agreement}
\end{table}

\paragraph{Instruction Data Conversion.}

To support instruction-based evaluation, we convert each annotated instance into a structured instruction-style format using expert-crafted task-specific prompt as outlined below.

\paragraph{Evaluation Metric.}

We formulate intent recognition as a multi-class classification problem and evaluate model performance using Accuracy~\citep{makridakis1993accuracy} (Appendix~\ref{sec:app_metrics}). Accuracy measures the proportion of instances for which the predicted intent label exactly matches the annotated ground-truth label, providing a direct assessment of overall classification correctness.

\begin{tcolorbox}[
  colback=lightblue!10,
  colframe=jpblue!76,
  title={\small Task Instruction for \textit{JF-ICR} },
  fontupper=\small,
  left=1mm,
  right=1mm,
  top=1mm,
  bottom=1mm
]
\begin{verbatim}
You are a Japanese financial expert fluent in 
Japanese business communication. Given a
financial question and the corresponding 
company response (both in Japanese), your task 
is to determine the company’s underlying intent 
level and output exactly one label from the 
following set: {"+2", "+1", "0", "-1", "-2"}.

The label meanings are provided for explanation 
only (DO NOT output these texts):
  "+2" : "Strong Commitment"
  "+1" : "Weak or Qualified Commitment"
  "0"  : "Neutral or Hedged Intent"
  "-1" : "Weak Refusal"
  "-2" : "Strong Refusal"

Financial Question: {Japanese question}
Company Response: {Japanese answer}

Directly output the chosen label, and do not 
provide any explanation.
Answer: {Intent label}
\end{verbatim}
\end{tcolorbox}

\subsection{\textit{JF-TE}: Japanese Financial Term Extraction}

Japanese financial terminology is shaped by the interaction of three writing systems, nested nominal compounds, and extensive borrowing from both Chinese and English. Financial terms are predominantly realized in kanji and katakana, where loanwords frequently undergo semantic drift from their source languages, and visually identical forms may convey entirely different meanings across languages (e.g., \emph{kentō} “consideration”, \emph{bimyō} “ambiguous”, \emph{taiō} “response”), limiting the effectiveness of cross-lingual dictionary-based approaches \citep{goworek2025bridging, li2020simple}. Beyond linguistic factors, Japanese financial disclosure follows conventions that diverge substantially from U.S.-style reporting: although disclosures are nominally investor-facing, they are primarily oriented toward shareholders, resulting in distinctive writing styles and information structuring. Annual filings are highly standardized, with meaningful updates often embedded in \textit{seemingly peripheral note sections} rather than headline figures, making them easy for language models to overlook. The \textit{JF-TE} task therefore targets financial term extraction from Japanese disclosure notes, assessing how well LLMs predominantly trained on U.S. financial documents can identify domain-specific terminology in Japanese financial contexts.

\paragraph{Task Definition}

The \textit{JF-TE} task is formulated as a financial terminology extraction and ranking problem over Japanese financial text, with explicit modeling of nested term structures. Inspired by termhood-oriented principles such as C-value~\citep{frantzi2000cvalue}, the task evaluates whether models can identify and prioritize domain-specific financial terminology in Japanese disclosures.
Each input is a single note section \( n \) drawn from professional Japanese financial documents. The model is required to identify a set of \emph{maximal financial terms} \( M = \{ m_1, m_2, \ldots, m_p \} \), where each \( m_i \) denotes the longest contiguous expression representing a financial concept in \( n \).
For each maximal term \( m_i \), the model further identifies a set of candidate nested terms \( T_i = \{ t_{i1}, t_{i2}, \ldots, t_{ik} \} \), where each \( t_{ij} \subseteq m_i \) is a potential financial term (including \( t_{ij} = m_i \)); in the absence of nested terms, \( T_i = \{ m_i \} \). The model outputs an ordered list \( T_i' \), ranking these candidates from the most to the least likely to represent domain-specific financial terminology. Non-financial expressions are excluded from the output.
The final output is a structured JSON object aggregating all \( T_i' \) across maximal terms. 

\paragraph{Data Source.}

To construct the \textit{JF-TE} dataset, we collect Japanese Annual Securities Reports (\textit{Y\=uka sh\=oken h\=okokusho}) disclosed through EDINET\footnote{\tiny \url{https://disclosure2.edinet-fsa.go.jp/}}, the official electronic disclosure system operated by Japan’s Financial Services Agency. Our corpus consists of 10 professional disclosures released in 2025 by 10 publicly listed companies, with each report spanning up to 261 pages in the original EDINET filings.

Consistent with Japanese financial disclosure practices, substantive updates and domain-specific information are often embedded in fine-grained appendix notes rather than the main narrative sections. To capture these information-dense units, we manually identify and extract individual note entries from each financial disclosure, treating each note as a standalone instance for expert annotation and financial terminology extraction. In total, this process yields 202 curated note-level data instances.

\paragraph{Expert Annotation.}

We follow a similar expert annotation protocol as in \textit{JF-ICR}, with task-specific guidelines tailored to hierarchical financial terminology (Appendix~\ref{sec:app_te_ag}). Each note-level instance was annotated by identifying the longest financial term spans and, where applicable, the nested financial terms contained within them. Annotation was conducted at the span level, allowing multi-word expressions, compound nouns, and nested structures to be marked explicitly.
The annotation guidelines were iteratively refined through pre-annotation rounds, with explicit rules governing term boundary decisions and the distinction between financial terminology and general expressions. All annotations were carried out by the same native-level Japanese financial experts (Appendix~\ref{sec:app_demography}) using the Label Studio platform (Figure~\ref{fig_labelstudio_te}, Appendix~\ref{sec:app_labelstudio}), ensuring consistency and reproducibility.

\paragraph{Quality Validation.}

We evaluate annotation quality using inter-annotator agreement metrics 
Macro-F1~\citep{sokolova2009systematic}, Cohen’s $\kappa$~\citep{cohen1960kappa}, and Krippendorff’s $\alpha$~\citep{krippendorff2011alpha} (Appendix~\ref{sec:app_validation}). 
The results demonstrate strong agreement in identifying financial term spans (Table~\ref{tab:annotation-agreement}). This consistency indicates that the annotation rules governing term boundaries and financial termhood are operational and reproducible, supporting the reliability of the resulting high-quality JF-TE dataset.

\paragraph{Instruction Data Conversion.}

To support instruction-based evaluation, we reformulate each annotated instance into an instruction-style format, with task-specific prompts tailored to financial terminology extraction as shown below.

\paragraph{Evaluation Metrics.}

The \textit{JF-TE} task is evaluated using a two-level metric design to reflect the hierarchical structure of financial terminology in Japanese disclosures. We first evaluate maximal financial term extraction by comparing predicted ones with the ground truth using exact matching and report the F1 score~\citep{tjong-kim-sang-de-meulder-2003-introduction} (Appendix~\ref{sec:app_metrics}). We then assess the ranking of nested financial terms within each maximal financial term using HitRate@K~\citep{bordes2013transe} (Appendix~\ref{sec:app_metrics}), which measures whether gold nested terms appear among the top-$K$ predictions. To account for variation in the number of nested terms across instances, we report HitRate@1, HitRate@5, and HitRate@10. Together, these metrics capture both accurate identification of complete financial term boundaries and effective prioritization of nested financial terminology.


\begin{tcolorbox}[
  colback=lightblue!10,
  colframe=jpblue!76,
  title={\small English Translated Task Instruction for \textit{JF-TE} },
  fontupper=\small,
  left=1mm,
  right=1mm,
  top=1mm,
  bottom=1mm
]
\begin{verbatim}
You are a Japanese financial expert specializing
in financial disclosure analysis. Read the 
following Japanese financial note carefully.

Identify all maximal (longest) financial terms 
in the text. For each maximal term, identify any 
nested financial terms it contains, including 
the maximal term itself if applicable.

For each maximal financial term, output a ranked 
list of its associated financial terms 
(including itself and any nested terms), ordered 
from most to least likely to represent domain-
specific financial terminology. Do not include 
non-financial expressions.

Return the output in JSON format, where each 
element corresponds to one maximal financial 
term and contains a ranked list of financial 
terms.

Text: {Japanese financial disclosure note}
Answer: {JSON with ranked financial terms}
\end{verbatim}
\end{tcolorbox}

\section{Experimental Results}

\subsection{Evaluation Models}

We evaluate a comprehensive set of 22 LLMs on the \textsc{\ebisu} benchmark, spanning a wide range of model types and training paradigms. The evaluated models comprise: 
\textbf{(1) 11 open-source general-purpose models} at both large and small scales 
(Llama-4-Scout-17B~\citep{meta_llama4_blog}, 
Llama-3.3-70B-Instruct~\citep{llama3_herd}, 
Qwen3-235B-A22B-Instruct-2507-FP8, 
Qwen3-32B, 
Qwen3-14B, 
Qwen3-8B~\citep{qwen3technicalreport}, 
Qwen2.5-32B-Instruct~\citep{qwen25technicalreport}, 
Qwen-14B~\citep{qwen_techreport}, 
Ministral-3-14B-Instruct-2512~\citep{mistral3_release}, 
DeepSeek-V3.1~\citep{deepseekai2024deepseekv3technicalreport}, 
Kimi-K2-Instruct~\citep{kimi_k2_techreport}); 
\textbf{(2) 4 proprietary closed-source models} accessed via official APIs 
(gpt-5~\citep{openai_gpt5_system_card_2025}, 
gpt-4o~\citep{openai_gpt4o_system_card_2024}, 
gemini-3-flash~\citep{gemini3_flash_modelcard}, 
claude-sonnet-4-5~\citep{claude_sonnet45_system_card}); 
\textbf{(3) English financial domain-specific model} 
(FinMA-7B~\citep{xie2023pixiu}; 
\textbf{(4) 5 Japanese general-purpose models} 
(Llama-3.3-Swallow-70B-Instruct-v0.4~\citep{swallow2025llama33swallow70b_instruct_v04}, 
TinySwallow-1.5B-Instruct~\citep{shing2025taid}, 
japanese-stablelm-instruct-beta-70b, 
japanese-stablelm-instruct-beta-7b~\citep{stability_japanese_stablelm_beta_2023}, 
nekomata-14b~\citep{sawada2024release_japanese_pretrained}); 
and \textbf{(5) Japanese financial domain-specific model} 
(nekomata-14b-pfn-qfin~\citep{hirano2024domain}). 
 

\subsection{Implementation Details}

To ensure evaluation integrity and consistency, we develop a unified evaluation pipeline based on the LM Evaluation Harness~\citep{eval-harness}. 
Proprietary models are evaluated using their official APIs, with temperature fixed to 0 to ensure deterministic outputs.. Open-source models available through TogetherAI are served and evaluated via the TogetherAI platform, and all other open-source models are deployed and evaluated locally using vLLM~\citep{kwon2023efficient} on GPUs. In-house evaluation is conducted on a cluster of four H100 GPUs, each equipped with 80GB of memory, for a total of over 200 GPU-hours.
We standardize the maximum generation length to 1,024.

\begin{table}[t]
\renewcommand{\arraystretch}{1}
\scriptsize
\centering
\scalebox{0.73}{
\begin{tabular}{@{}lcccccc@{}}
\toprule
\textbf{Model} & \textbf{\textit{JF-ICR}} & \multicolumn{4}{c}{\textbf{\textit{JF-TE}}} & \textbf{Mean} \\
\cmidrule(lr){2-2}\cmidrule(lr){3-6}\cmidrule(lr){7-7}
& \textbf{Acc} & \textbf{F1} & \textbf{HR@1} & \textbf{HR@5} & \textbf{HR@10} & \textbf{Avg} \\
\midrule

\rowcolor{lightblue}
\multicolumn{7}{c}{\textit{\textbf{Open-source General Models}}} \\
Llama-4-Scout-17B & \textbf{0.6064} & 0.4133 & 0.1085 & 0.3094 & 0.4513 & 0.3778 \\
Llama-3.3-70B-Instruct & 0.5638 & \underline{0.4370} & \textbf{0.1277} & \textbf{0.3657} & \textbf{0.5111} & \textbf{0.4011} \\
Qwen3-235B-A22B-Instruct & 0.4574 & 0.4355 & 0.0971 & 0.3053 & 0.4301 & 0.3451 \\
Qwen3-32B & 0.1915 & 0.2860 & 0.0285 & 0.1052 & 0.1763 & 0.1575 \\
Qwen3-14B & 0.1809 & 0.2038 & 0.0165 & 0.0787 & 0.1097 & 0.1179 \\
Qwen3-8B & 0.1277 & 0.0548 & 0.0050 & 0.0142 & 0.0233 & 0.0450 \\
Qwen2.5-32B-Instruct & 0.3830 & 0.4137 & 0.0982 & \underline{0.3462} & \underline{0.4786} & 0.3439 \\
Qwen-14B & 0.0638 & 0.1613 & 0.0308 & 0.0650 & 0.1076 & 0.0857 \\
Ministral-3-14B-Instruct & 0.3723 & 0.3756 & 0.0875 & 0.2377 & 0.3704 & 0.2887 \\
DeepSeek-V3.1 & 0.5106 & 0.3653 & 0.0794 & 0.2663 & 0.4121 & 0.3267 \\
Kimi-K2-Instruct & 0.4574 & \textbf{0.4396} & 0.1048 & 0.3251 & 0.4680 & 0.3590 \\

\midrule
\rowcolor{lightblue}
\multicolumn{7}{c}{\textit{\textbf{Proprietary Models}}} \\
gpt-5 & 0.5532 & 0.0000 & 0.0000 & 0.0000 & 0.0000 & 0.1106 \\
gpt-4o & 0.0319 & 0.3997 & 0.0890 & 0.2845 & 0.4255 & 0.2461 \\
gemini-3-flash & 0.3936 & 0.0000 & 0.0000 & 0.0000 & 0.0000 & 0.0787 \\
claude-sonnet-4-5 & 0.5213 & 0.4362 & 0.0743 & 0.3022 & 0.4752 & 0.3618 \\

\midrule
\rowcolor{lightblue}
\multicolumn{7}{c}{\textit{\textbf{English Financial Models}}} \\
FinMA-7B & 0.1596 & 0.0000 & 0.0000 & 0.0000 & 0.0000 & 0.0319 \\

\midrule
\rowcolor{lightblue}
\multicolumn{7}{c}{\textit{\textbf{Japanese General Models}}} \\
Llama-3.3-Swallow-70B-Instruct & \underline{0.5957} & 0.4070 & \underline{0.1258} & 0.3270 & 0.4701 & \underline{0.3851} \\
TinySwallow-1.5B-Instruct & 0.0638 & 0.1937 & 0.0264 & 0.0712 & 0.1012 & 0.0913 \\
japanese-stablelm-instruct-beta-7b & 0.0426 & 0.0143 & 0.0069 & 0.0079 & 0.0079 & 0.0159 \\
japanese-stablelm-instruct-beta-70b & 0.0319 & 0.1826 & 0.0388 & 0.1090 & 0.1461 & 0.1017 \\
nekomata-14b & 0.0213 & 0.0059 & 0.0173 & 0.0330 & 0.0479 & 0.0251 \\

\midrule
\rowcolor{lightblue}
\multicolumn{7}{c}{\textit{\textbf{Japanese Financial Model}}} \\
nekomata-14b-pfn-qfin & 0.0213 & 0.0468 & 0.0012 & 0.0118 & 0.0181 & 0.0198 \\

\bottomrule
\end{tabular}}
\caption{LLM performance on the \textsc{\ebisu} benchmark. \textit{JF-ICR} is evaluated using Accuracy (Acc). \textit{JF-TE} is evaluated using F1 and HitRate@K (HR@1/5/10). Bold values denote the best scores and underlined values denote the second-best scores in each column.}
\label{tab:ebisu-results}
\end{table}

\begin{figure}[ht]
  \includegraphics[width=1.0\linewidth]{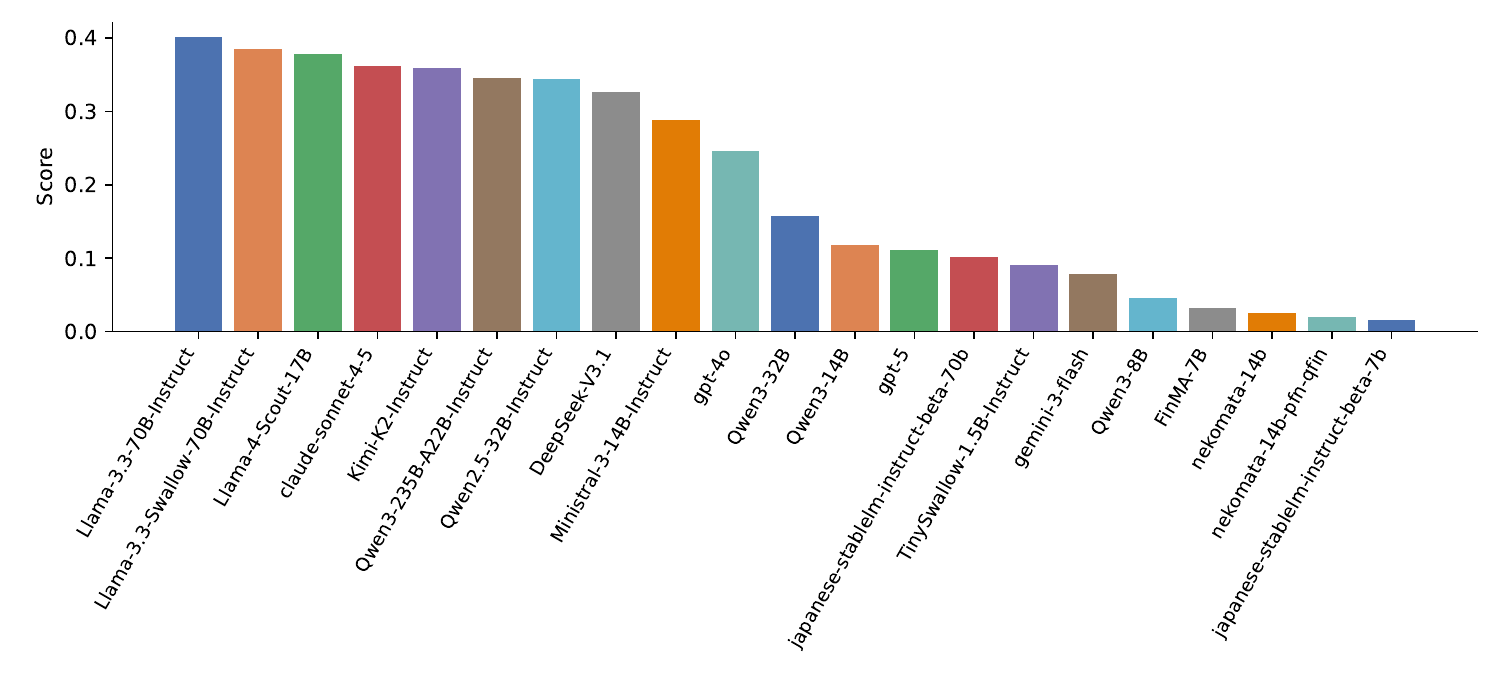}
  \caption {Ranked models performance on the \textsc{\ebisu} benchmark..}
  \label{fig_results}
\end{figure}

\subsection{Key Results}

Table~\ref{tab:ebisu-results} and Figure~\ref{fig_results} summarize model performance on the \textsc{\ebisu} benchmark. 
We structure our analysis around the following research questions.

\paragraph{RQ1: Is \textsc{\ebisu} challenging for state-of-the-art LLMs under Japanese financial settings?}

\textit{Overall, the results show that current LLMs face substantial challenges on the \textsc{\ebisu} benchmark, including state-of-the-art systems.} Leading proprietary models such as GPT-4o (0.2461) and Claude-Sonnet-4.5 (0.3618) exhibit limited performance, while the best overall result is achieved by the open-source Llama-3.3-70B-Instruct (0.4011), suggesting persistent difficulties in robustly modeling the linguistic and pragmatic characteristics of Japanese financial texts.
\textit{Performance generally improves with increasing model scale.} For example, within the Qwen family, larger models such as Qwen3-235B-A22B-Instruct (0.3451) and Qwen3-32B (0.1575) outperform the smaller Qwen3-14B (0.1179) and Qwen3-8B (0.0450). However, these gains remain limited, indicating that the challenges reflected in \textsc{\ebisu} cannot be resolved through model capacity alone. Instead, the persistent performance gap points to linguistic and cultural properties of Japanese financial communication that are not sufficiently captured by prevailing training paradigms. 
Collectively, these results highlight the difficulty of Japanese financial language understanding and position \textsc{\ebisu} as a demanding benchmark for evaluating such capabilities in modern LLMs.

\paragraph{RQ2: How effectively do models transfer knowledge across language and domain boundaries in Japanese finance?}

Clear limitations emerge in cross-language and cross-domain knowledge transfer across model families. 
\textit{English financial domain-specific model (FinMA-7B: 0.0319) performs worse than general-purpose models of comparable scale like Qwen3-8B (0.0450)}. This pattern indicates that financial knowledge learned primarily from English corpora does not readily transfer to Japanese financial tasks without sufficient Japanese language training. 
\textit{Japanese general-purpose models still underperform relative to their English counterparts.} For instance, Llama-3.3-Swallow-70B-Instruct (0.3851) lags behind Llama-3.3-70B-Instruct (0.4011), indicating that performance gains from language adaptation do not necessarily generalizable.
\textit{In contrast to expectations, continued language- and domain-specific pretraining does not consistently improve performance.}
The Japanese financial model nekomata-14b-pfn-qfin (0.0198) performs worse than both its Japanese general counterpart nekomata-14b (0.0251) and the original backbone Qwen-14B (0.0857). This suggests that continued next-token pretraining on Japanese financial text alone is insufficient, and may even hinder performance on tasks requiring implicit intent understanding in Japanese finance.
Collectively, these findings show that \textsc{\ebisu} exposes critical shortcomings in existing English financial, Japanese general, and Japanese financial models, and provides a meaningful benchmark for driving progress toward models that better capture the linguistic and cultural context of Japanese finance.

\paragraph{RQ3: Do LLMs exhibit systematic biases when recognizing implicit commitment in Japanese financial responses?}

\textit{Performance on \textit{JF-ICR} falls short across all evaluated models, revealing persistent difficulties in intent understanding within Japanese financial contexts.} Even the strongest model, Llama-4-Scout-17B (Acc: 0.6064), struggles to accurately recognize implicit agreement and refusal in Japanese finance, indicating that this capability remains an open challenge.
Beyond overall accuracy, we further examine the average predicted commitment scores across models
. \textit{At similar model sizes, English-centric models consistently assign higher commitment scores than their Japanese counterparts}, such as Llama-3.3-70B-Instruct (0.9787) compared to Llama-3.3-Swallow-70B-Instruct (0.8404), and Qwen-14B (0.3830) compared to nekomata-14b (-0.8511). This pattern indicates a systematic bias in models primarily trained on English data, where indirect refusals and strategic non-commitment in Japanese financial responses are frequently misinterpreted as acceptance or weak agreement. 
These findings reflect the intrinsic difficulty of \textit{JF-ICR}, which arises from the combined influence of Japanese linguistic features that encode intent implicitly, such as sentence-final constructions and layered negation, and financial communication norms that discourage explicit refusal. As a result, \textit{JF-ICR} requires models to distinguish genuine commitment from strategic non-commitment beyond surface-level semantics in high-context Japanese financial discourse.

\paragraph{RQ4: Are Japanese financial terms particularly difficult for LLMs to extract due to language-specific lexical properties?}

\textit{Performance on \textit{JF-TE} remains weak across all evaluated models, revealing substantial limitations in current LLMs’ ability to extract and prioritize Japanese financial terminology.} Even the strongest-performing model, Llama-3.3-70B-Instruct, achieves only modest HitRate at different cutoffs (HitRate@1: 0.1277, HitRate@5: 0.3657, HitRate@10: 0.5111), indicating that accurate identification and ranking of financial terms in Japanese disclosures is far from solved. 
\textit{JF-TE} emphasizes that Japanese financial term extraction is not merely a problem of surface matching, but one of identifying and ranking domain-relevant terminology in a setting dominated by loanwords with shifted meanings and densely populated financial expressions shaped by Japanese disclosure conventions.

\section{Conclusion}

We present \textsc{\ebisu}, a benchmark for evaluating LLMs on Japanese financial language understanding, explicitly designed to reflect the linguistic structure and communication norms of Japanese finance. Comprising \textit{JF-ICR} and \textit{JF-TE}, \textsc{\ebisu} targets two core challenges that arise from the interaction between Japanese language use and Japanese financial practices: implicit commitment recognition and financial term extraction. Experiments across diverse model families show that current LLMs, including state-of-the-art proprietary and open-source systems, struggle on both tasks. While increased model scale yields some improvements, language- and domain-specific training provides limited gains and can even degrade performance. These results reveal persistent gaps rooted in the linguistic and cultural characteristics of Japanese financial communication not solved by current model trainings. By exposing systematic difficulties in commitment recognition and terminology ranking, \textsc{\ebisu} offers a focused benchmark for advancing LLMs toward deeper and more reliable understanding of Japanese finance.

\section*{Limitations}

Several limitations of \textsc{\ebisu} warrant acknowledgment. First,the dataset scope is constrained by practical considerations: \textit{JF-ICR} comprises Q\&A transcripts from 4 companies over a limited temporal window (2023--2026), while \textit{JF-TE} focuses exclusively on Annual Securities Reports from EDINET. This coverage may not fully represent the diversity of Japanese financial communication across company sizes, industries, or disclosure formats. Second, due to the current dataset size, we do not provide train/test splits in this release. Once the dataset is expanded, standard train/validation/test splits will be introduced to support model training and evaluation. Third, \textsc{\ebisu} evaluates two complementary tasks that target core challenges in Japanese financial language understanding but do not encompass the full spectrum of financial NLP capabilities. Both tasks operate on relatively short text segments, limiting evaluation of long-context understanding and multi-turn discourse reasoning. Finally, the evaluation metrics have inherent limitations: Accuracy for \textit{JF-ICR} treats all misclassifications equally, while exact matching for \textit{JF-TE} may penalize semantically correct but slightly misaligned spans.
    

\section*{Ethical Concerns}

The release of \textsc{\ebisu} raises several ethical considerations that warrant careful attention. A primary concern is a risk of over-reliance on benchmark scores for evaluating Japanese financial LLM capabilities. Benchmark performance may not fully predict real-world utility in production systems, where factors such as latency, robustness to distribution shift, and integration with existing workflows are equally critical. The focused scope of \textsc{\ebisu} introduces potential dataset bias and representativeness limitations. Because the datasets are drawn from a limited set of large, publicly listed companies, they may not reflect communication styles, terminology, or disclosure practices across the broader Japanese financial landscape. The temporal scope (2023--2026) may also not capture evolving financial terminology or regulatory changes. Moreover, as with many benchmarks, there is a risk that models may be optimized to perform well on \textsc{\ebisu} without corresponding improvements in generalization, potentially encouraging benchmark-specific overfitting.

\textsc{\ebisu} is released for research and evaluation purposes to advance the understanding of Japanese financial language processing capabilities in large language models. All datasets are constructed from publicly available corporate disclosures and financial documents, and we respect the original data sources' terms of use and licensing requirements.The benchmark is intended for academic research, model evaluation, and methodological development. Models that achieve high performance on \textsc{\ebisu} should not be assumed suitable for production deployment in financial systems, automated investment analysis, regulatory compliance, or other high-stakes applications without extensive additional validation, human oversight, and domain expert review. The tasks evaluated in \textsc{\ebisu} focus on implicit commitment recognition and financial term extraction, which represent specific aspects of Japanese financial language understanding but do not guarantee accurate financial reasoning, risk assessment, or regulatory interpretation. We encourage responsible use of the benchmark and welcome feedback from the research community to improve its coverage, fairness, and utility.


\bibliography{custom}

\appendix
\newpage



\section{\textit{JF-ICR} Annotation Guideline}
\label{sec:app_icr_ag}
\paragraph{Annotation Philosophy.}

\textit{JF-ICR} focuses on the degree of commitment expressed in management responses, rather than the surface intent of the question or the factual correctness of the content. Annotations are determined solely by whether and how the answer conveys a forward-looking action, decision, or policy stance, reflecting common communication practices in financial disclosures.

Due to the cautious and convention-driven nature of Japanese financial disclosures, explicit strong commitments are relatively rare.
As a result, a substantial portion of responses naturally fall into weakly committed or neutral categories, reflecting realistic managerial communication practices rather than annotation ambiguity or inconsistency.

\subsection{Classification Categories}
Each answer is annotated based on the speaker’s degree of commitment toward a future action, decision, or policy stance expressed in the response.

\begin{itemize}
    \item \textit{+2: Strong Commitment}--clear, explicit, and unambiguous commitment to a concrete action or outcome.
    \item \textit{+1: Weak or Qualified Commitment}--intention or directional stance expressed with conditions, hedging, or limited scope.
    \item \textit{0: Neutral or Hedged Intent}--clarification, explanation, assessment, or perception without commitment or refusal.
    \item \textit{-1: Weak Refusal}--implicit or qualified rejection of a proposed action.
    \item \textit{-2: Strong Refusal}--explicit and definitive rejection with no flexibility.
\end{itemize}

\subsection{General Annotation Rules}
\begin{itemize}
    \item Annotation focuses on \textbf{intent and commitment strength}, not factual correctness.
    \item Only statements expressing \textbf{future actions, decisions, policies, targets, or strategic directions} are considered for non-zero labels.
    \item Annotation is based on the \textbf{content of the answer}, not the surface form or intent of the question.
    Even if the question is retrospective or explanatory, the answer may introduce forward-looking commitment or refusal.
    \item If an answer does not express any future-oriented action, decision, or policy stance, it should be annotated as \textit{0}, regardless of the question.
    \item Hedging expressions (e.g., uncertainty, dependency on conditions, ongoing review) reduce commitment strength.
    \item When multiple signals appear, the label reflects the \textbf{strongest applicable commitment or refusal} expressed in the answer.
    \item Annotation is based on the \textbf{overall stance of the full response}, rather than isolated sentences or individual expressions; labels reflect the speaker’s dominant commitment or refusal as conveyed by the answer as a whole.

    \item Annotations focus on the \textbf{strength of commitment}, rather than the tone,
    length, or politeness of the response.
    \item Conventional polite expressions in Japanese financial discourse
    (e.g., \small\begin{CJK}{UTF8}{min}「努めてまいります」\end{CJK})
    do not by themselves imply commitment.

\end{itemize}

\subsection{Specific Annotation Rules}
\begin{itemize}
\item \textit{+2 (Strong Commitment):}  
Applied when the answer contains a clear, decisive statement indicating a firm commitment.
Typical linguistic signals include:
\begin{itemize}
    \item \small\begin{CJK}{UTF8}{min}「〜します」「〜を実施します」「〜を達成します」\end{CJK}
    \item \small\begin{CJK}{UTF8}{min}「〜を決定しています」「〜は確定しています」\end{CJK}
    \item \small\begin{CJK}{UTF8}{min}「方針を変更する考えはありません」\end{CJK}
    \quad (when directly answering a decision or policy question)
\end{itemize}
Representative examples include:
\begin{itemize}
    \item \small\begin{CJK}{UTF8}{min}「来期は増配を実施します。」\end{CJK}
    \item \small\begin{CJK}{UTF8}{min}「中長期でROE12\%を達成します。」\end{CJK}
    \item \small\begin{CJK}{UTF8}{min}「この施策により利益成長は実現できると確信しています。」\end{CJK}
\end{itemize}

\item \textit{+1 (Weak or Qualified Commitment):}  
Applied when the answer expresses a positive or leaning commitment toward a future action or outcome,
but with qualifications, caution, or limited specificity.
Such responses indicate directional intent or possibility rather than a finalized decision.
Typical linguistic signals include:
\begin{itemize}
    \item \small\begin{CJK}{UTF8}{min}「〜していきたい」「〜を目指しています」\end{CJK}
    \item \small\begin{CJK}{UTF8}{min}「〜と考えています」「〜を見込んでいます」\end{CJK}
    \item Conditional or hypothetical expressions (e.g., \small\begin{CJK}{UTF8}{min}「〜であれば」「〜次第で」\end{CJK})
\end{itemize}
Representative examples include:
\begin{itemize}
    \item \small\begin{CJK}{UTF8}{min}「成長投資を進めていきたいと考えています。」\end{CJK}
    \item \small\begin{CJK}{UTF8}{min}「今後も収益は拡大していくと見ています。」\end{CJK}
    \item \small\begin{CJK}{UTF8}{min}「環境が整えば、検討を進める考えです。」\end{CJK}
\end{itemize}

\item \textit{0 (Neutral or Hedged Intent):}  
Applied when the answer exhibits genuine ambiguity or provides clarification, explanation, or background
without expressing commitment or refusal toward any future action, decision, or policy stance.
Such responses may describe facts, context, perceptions, or reasoning, but do not convey a directional
or evaluative position regarding future behavior.
Typical linguistic signals include:
\begin{itemize}
    \item \small\begin{CJK}{UTF8}{min}「〜断定できません」「明確な見通しは示せない」\end{CJK}
    \item \small\begin{CJK}{UTF8}{min}「検討中」「状況を見極める必要がある」\end{CJK}
    \item Purely descriptive or explanatory statements providing facts or background
\end{itemize}
Representative examples include:
\begin{itemize}
    \item \small\begin{CJK}{UTF8}{min}「現時点では明確な見通しは示せません。」\end{CJK}
    \item \small\begin{CJK}{UTF8}{min}「様々な見方があり、コメントは差し控えます。」\end{CJK}
    \item \small\begin{CJK}{UTF8}{min}「過去にはこのような取り組みを行ってきました。」\end{CJK}
    \quad (background explanation only)
\end{itemize}

\item \textit{-1 (Weak Refusal):}  
Applied when the answer expresses a negative stance toward a proposed future action or change,
but does so in a qualified, conditional, or time-bound manner that leaves room for future reconsideration.
Such responses reject the action in the current context without ruling it out permanently.
Typical linguistic signals include:
\begin{itemize}
    \item \small\begin{CJK}{UTF8}{min}「現時点では〜しない」「直ちには考えていない」\end{CJK}
    \item \small\begin{CJK}{UTF8}{min}「今後検討の余地はあるが」\end{CJK}
    \item Refusals framed as temporary, conditional, or dependent on future circumstances
\end{itemize}
Representative examples include:
\begin{itemize}
    \item \small\begin{CJK}{UTF8}{min}「現時点では配当方針を変更する考えはありません。」\end{CJK}
    \item \small\begin{CJK}{UTF8}{min}「今中計期間中に見直すことは想定していません。」\end{CJK}
    \item \small\begin{CJK}{UTF8}{min}「足元では難しいと考えていますが、今後は検討します。」\end{CJK}
\end{itemize}

\item \textit{-2 (Strong Refusal):}  
Applied when the answer clearly and definitively rejects a proposed future action, decision,
or policy stance, leaving no visible room for reconsideration.
Such responses convey a firm and final negative position.
Typical linguistic signals include:
\begin{itemize}
    \item \small\begin{CJK}{UTF8}{min}「〜する予定はありません」\end{CJK}
    \item \small\begin{CJK}{UTF8}{min}「〜は行いません」「〜を否定します」\end{CJK}
\end{itemize}
Representative examples include:
\begin{itemize}
    \item \small\begin{CJK}{UTF8}{min}「株式分割を行う予定はありません。」\end{CJK}
    \item \small\begin{CJK}{UTF8}{min}「当該事業への投資は実施しません。」\end{CJK}
    \item \small\begin{CJK}{UTF8}{min}「その想定は当社の方針ではありません。」\end{CJK}
\end{itemize}
\end{itemize}

\section{\textit{JF-TE} Annotation Guideline}
\label{sec:app_te_ag}
\paragraph{Annotation Philosophy.}
\textit{JF-TE} focuses on identifying financial and accounting terminology explicitly defined or referenced
in notes and annotations ({\small\begin{CJK}{UTF8}{min}注記・※\end{CJK}}) 
of Japanese Annual Securities Reports.
These sections are primarily functional and explanatory, serving to define accounting concepts,
clarify disclosure scope, or specify calculation and regulatory conditions, rather than to provide
narrative or managerial discussion.

Annotations aim to capture domain-relevant financial terms as they appear in disclosure notes,
reflecting realistic accounting and reporting practices.

\subsection{Scope of the Text}
All input texts are extracted exclusively from non-narrative explanatory sections, including:
\begin{itemize}
    \item Footnotes and annotation blocks ({\small\begin{CJK}{UTF8}{min}注記・※\end{CJK}})
    \item Explanatory remarks below tables
    \item Definitions of accounting items, metrics, or disclosure conditions
    \item Clarifications of calculation methods, scope, or regulatory treatment
\end{itemize}

These texts should be interpreted within their definitional and regulatory context.

\subsection{Annotation Unit}
Each annotation unit corresponds to a single text span (NER-style span).
Annotators should:
\begin{itemize}
    \item Select only the \textbf{minimal span} that represents a financial or accounting term
    \item Prefer noun or noun-phrase spans
    \item Avoid including surrounding explanatory text, verbs, or modifiers
\end{itemize}

If a span does not represent a financial or accounting concept, it should not be annotated.

\subsection{Definition of Financial Terms}
A financial term is defined as a word or phrase that denotes:
\begin{itemize}
    \item An accounting item or financial statement component
    \item A financial metric, ratio, or valuation concept
    \item A disclosure category, reporting unit, or consolidation scope
    \item A legally or regulatorily defined financial or accounting concept
\end{itemize}

General language or procedural expressions should not be annotated unless they constitute
recognized financial terminology.

\subsection{Nested and Overlapping Terms}
Financial terms may appear in nested or overlapping forms.
Annotators should follow these rules:
\begin{itemize}
    \item Both nested and non-nested financial terms should be annotated if they independently
    represent meaningful financial or accounting concepts.
    \item When a longer term contains a shorter term, each should be annotated separately,
    provided that each span has standalone financial meaning.
    \item Annotators should always select the \textbf{minimal span} for each term,
    even when multiple annotated spans overlap.
    \item Nested terms should not be merged unless the shorter term lacks independent
    financial meaning outside the longer expression.
\end{itemize}

\subsection{Exclusions}
The following should \textbf{not} be annotated:
\begin{itemize}
    \item General verbs, adjectives, or explanatory phrases
    \item Purely numerical values or units (unless part of a named financial term)
    \item Time expressions without financial meaning
    \item Cross-references, footnote symbols, or formatting markers
\end{itemize}

\subsection{Representative Examples}

\paragraph{Example 1: Single Financial Term.}
\begin{quote}
\small\begin{CJK}{UTF8}{min}（注）当社は、金融危機等の状況下でも安定した資金確保を目的として、取引銀行とコミットメントラインを設定しました。この契約に基づく借入未実行残高は次のとおりであります。
\end{CJK}
\end{quote}

Annotated spans:
\begin{itemize}
    \item \small\begin{CJK}{UTF8}{min}取引銀行\end{CJK}
    \item \small\begin{CJK}{UTF8}{min}コミットメントライン\end{CJK}
    \item \small\begin{CJK}{UTF8}{min}借入未実行残高\end{CJK}
\end{itemize}
No nested annotation is applied in this example, as the identified financial terms
represent parallel disclosure units rather than compositional or hierarchical expressions.

\paragraph{Example 2: Nested Financial Terms.}
\begin{quote}
\small\begin{CJK}{UTF8}{min}
(※) 潜在株式調整後1株当たり当期純利益については、潜在株式がないため記載しておりません。
\end{CJK}
\end{quote}

Annotated spans:
\begin{itemize}
    \item \small\begin{CJK}{UTF8}{min}潜在株式調整後1株当たり当期純利益\end{CJK}
    \item \small\begin{CJK}{UTF8}{min}潜在株式\end{CJK}
    \item \small\begin{CJK}{UTF8}{min}1株当たり当期純利益\end{CJK}
    \item \small\begin{CJK}{UTF8}{min}当期純利益\end{CJK}
    \item \small\begin{CJK}{UTF8}{min}純利益\end{CJK}
\end{itemize}

\paragraph{Example 3: Non-Annotated Content.}
\begin{quote}
\small\begin{CJK}{UTF8}{min}
(注) 主に通信販売している機能性表示食品「ごま豆乳仕立てのみんなのみかたDHA」、特定保健用食品「イマークＳ」などの健康食品。
\end{CJK}
\end{quote}

No annotation is applied, as the sentence does not introduce a specific financial or accounting term.

\section{Annotator Demography}
\label{sec:app_demography}

All annotations in \textsc{\ebisu} were conducted by annotators with strong financial expertise, substantial professional experience, and high proficiency in Japanese, ensuring both domain fidelity and annotation reliability.

One annotator is a doctoral student in Japan with a solid academic background in financial mathematics and several years of study and residence in Japan. Their research focuses on AI applications in finance, and they have previously contributed to the annotation of financial benchmarks targeting financial reasoning tasks. Prior to their doctoral studies, they accumulated professional experience as a quantitative researcher in the financial industry, providing practical exposure to real-world financial data and analytical workflows.

Another annotator is a researcher at a Japanese fintech company with nearly two decades of experience in the Japanese financial market. They hold a Ph.D. in Economics and possess deep expertise in Japanese corporate finance, financial reporting, and investor communications. As a native-level Japanese speaker, they are familiar with the linguistic, cultural, and institutional conventions of Japanese financial disclosures, including earnings reports, management commentary, and shareholder meeting materials.

Together, the annotators’ combined academic training, industry experience, and linguistic competence enable the construction of a high-quality Japanese financial benchmark. Their expertise ensures that the annotations are both technically accurate and contextually grounded, providing a reliable foundation for evaluating Japanese financial language understanding.

\newpage
\onecolumn
\section{Annotation Process}
\label{sec:app_labelstudio}

\begin{figure*}[ht]
  \includegraphics[width=0.9\linewidth]{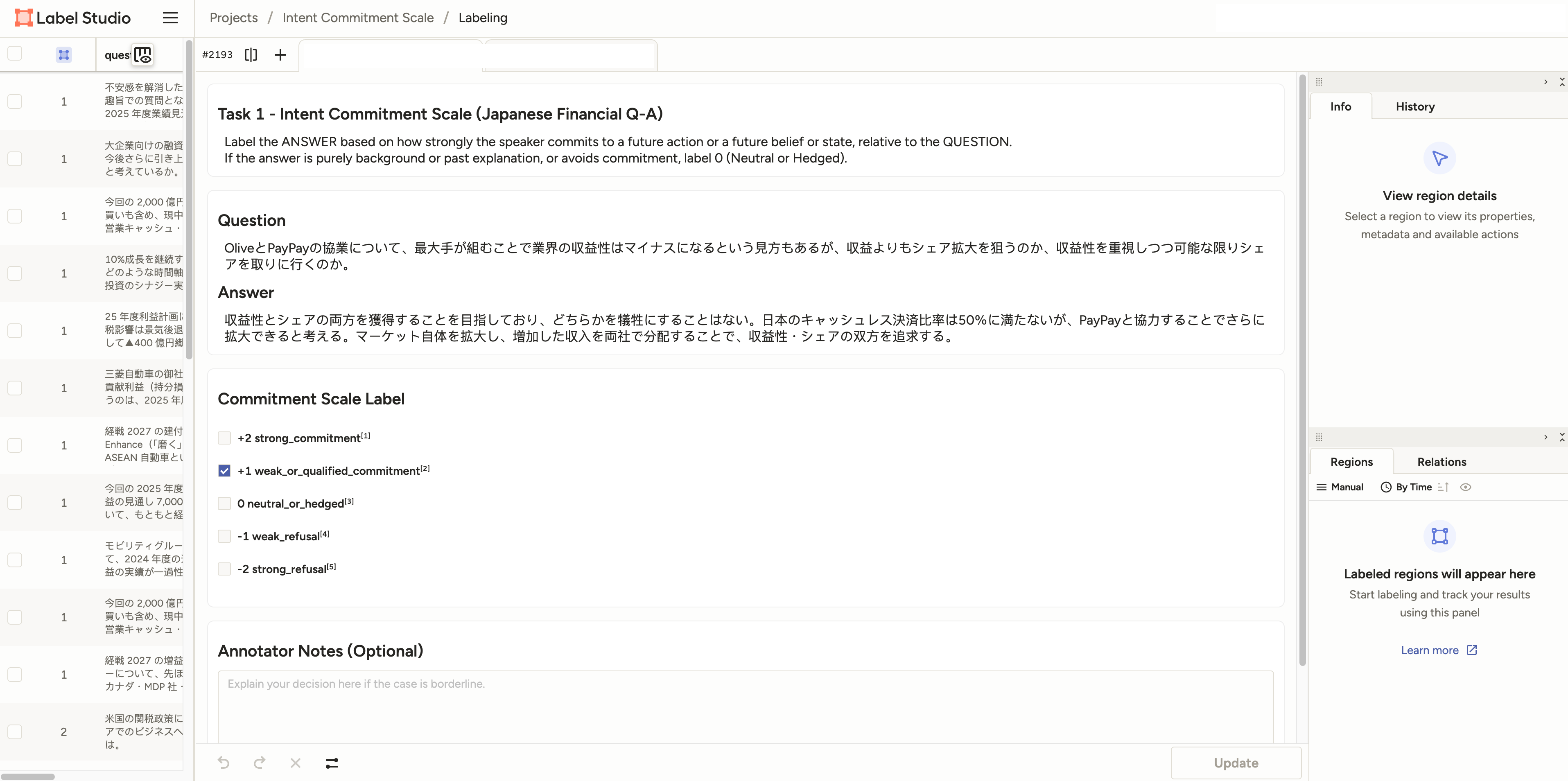}
  \caption {The Label Studio interface of the \textit{JF-ICR} annotation process.}
  \label{fig_labelstudio_icr}
\end{figure*}

\begin{figure*}[ht]
  \includegraphics[width=0.9\linewidth]{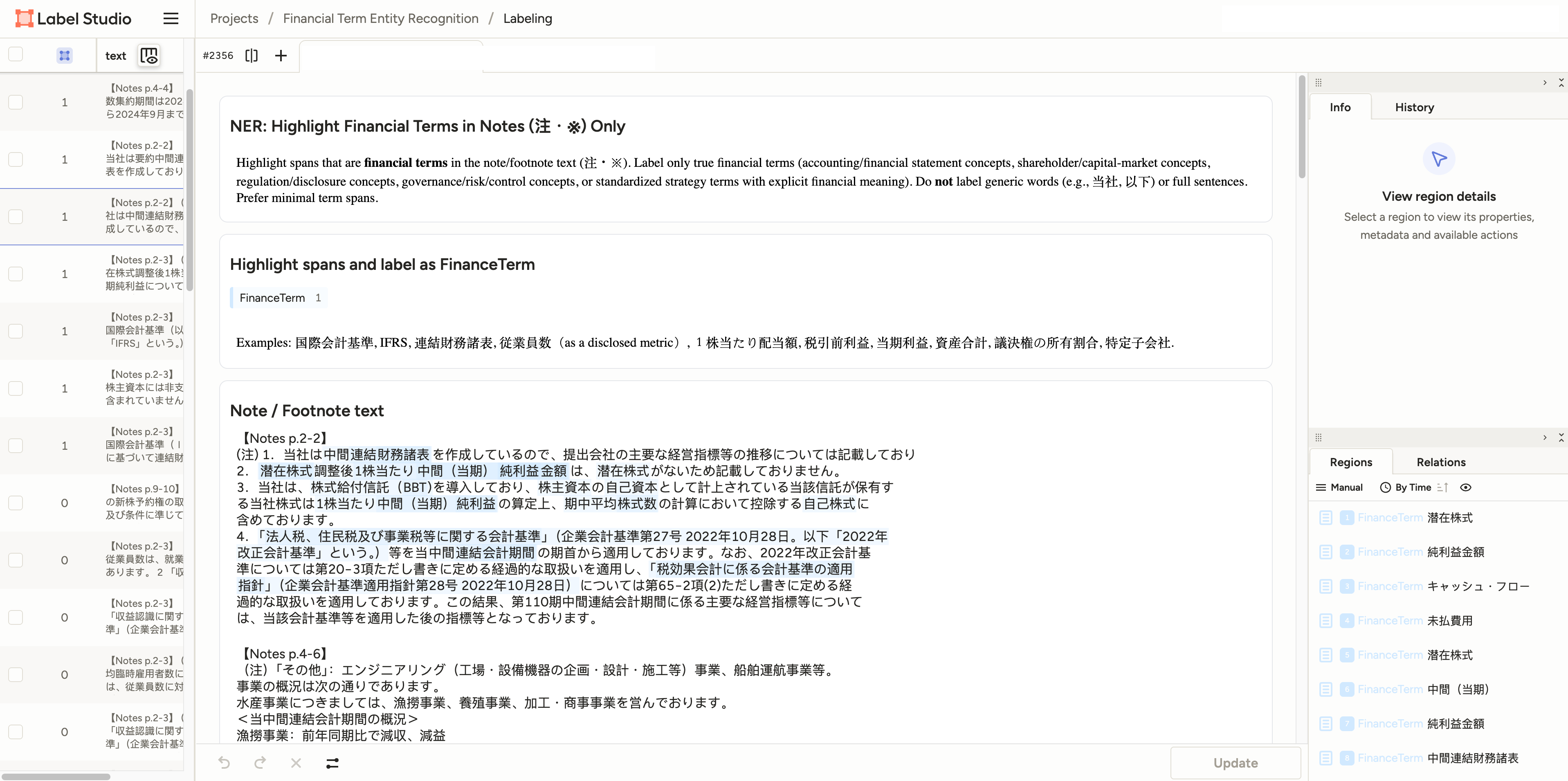}
  \caption {The Label Studio interface of the \textit{JF-TE} annotation process.}
  \label{fig_labelstudio_te}
\end{figure*}

\newpage
\twocolumn
\section{Quality Validation}
\label{sec:app_validation}


To assess the reliability of the \textit{JF-ICR} and \textit{JF-TE} annotations, we compute 3 standard \textbf{inter-annotator agreement metrics}:
Macro-F1, Cohen’s $\kappa$, and Krippendorff’s $\alpha$.
Below we summarize their formal definitions.

\paragraph{Macro-F1.}

Let $\mathcal{C}$ denote the set of categories, and for each category
$c \in \mathcal{C}$, we compute
\begin{equation}
\mathrm{Precision}_c =
\frac{TP_c}{TP_c + FP_c},
\end{equation}

\begin{equation}
\mathrm{Recall}_c =
\frac{TP_c}{TP_c + FN_c}.
\end{equation}

The category-level F1 score is
\begin{equation}
\mathrm{F1}_c =
\frac{2 \cdot \mathrm{Precision}_c \cdot \mathrm{Recall}_c}
{\mathrm{Precision}_c + \mathrm{Recall}_c}.
\end{equation}

Macro-F1 averages the per-class F1 scores uniformly:
\begin{equation}
\mathrm{Macro\text{-}F1}
=
\frac{1}{|\mathcal{C}|}
\sum_{c \in \mathcal{C}}
\mathrm{F1}_c .
\end{equation}

Compared to accuracy, Macro-F1 weights all classes equally and is therefore
more sensitive to skewed label distributions.


\paragraph{Cohen’s $\kappa$.}

For two annotators $r_1$ and $r_2$, let
\begin{equation}
P_o = \sum_{c \in \mathcal{C}} p_{cc},
\end{equation}

\begin{equation}
P_e = \sum_{c \in \mathcal{C}} p_{c\cdot}\,p_{\cdot c},
\end{equation}
where $p_{cc}$ is the empirical probability that both annotators assign
category $c$, and $p_{c\cdot}$ and $p_{\cdot c}$ are the corresponding
marginal probabilities.

Cohen’s $\kappa$ corrects observed agreement by subtracting chance agreement:
\begin{equation}
\kappa =
\frac{P_o - P_e}{1 - P_e}.
\end{equation}

A value of $\kappa=1$ indicates perfect agreement, while $\kappa=0$
corresponds to chance-level agreement.

\paragraph{Krippendorff’s $\alpha$ (Nominal Scale).}

Krippendorff’s $\alpha$ accommodates multiple annotators and missing labels.
It is defined as
\begin{equation}
\alpha = 1 - \frac{D_o}{D_e}.
\end{equation}

For nominal categories, the distance function is
\begin{equation}
\delta(a,b) =
\begin{cases}
0, & a = b, \\
1, & a \neq b.
\end{cases}
\end{equation}

Let $N$ be the number of items, $n_i$ the number of annotations for item $i$,
$n_{ia}$ the number of times item $i$ receives label $a$, and
$n_a$ the total number of times label $a$ appears. Then
\begin{equation}
D_o =
\frac{1}{N}
\sum_{i=1}^{N}
\frac{1}{n_i - 1}
\sum_{a \neq b}
n_{ia}\,n_{ib},
\end{equation}

\begin{equation}
D_e =
\frac{1}{N(N-1)}
\sum_{a \neq b}
n_a\,n_b .
\end{equation}

When annotators completely agree, $\alpha = 1$, and larger values indicate
higher annotation reliability.

\section{Evaluation Metrics}
\label{sec:app_metrics}

The \textit{JF-ICR} task is an intent recognition problem, which is treated as a multi-class classification task over the
label set $\{+2,+1,0,-1,-2\}$.
Given a predicted label $\hat{y}_i$ and the annotated gold label $y_i$ for the $i$-th instance, \textbf{Accuracy} is defined as:
\begin{equation}
\mathrm{Acc}
=
\frac{1}{N}
\sum_{i=1}^{N}
\mathbf{1}\!\left(\hat{y}_i = y_i\right),
\end{equation}
where $\mathbf{1}(\cdot)$ is the indicator function and $N$ is the total number of instances.

The \textit{JF-TE} task is evaluated using a two-level metric design to reflect the hierarchical structure of financial terminology in Japanese disclosures.

\paragraph{Maximal Financial Term F1.}
For each instance, we derive the set of maximal financial terms from both gold
annotations ($M_{\text{gold}}$) and predictions ($M_{\text{pred}}$). We compute:
\begin{equation}
TP = \left| M_{\text{gold}} \cap M_{\text{pred}} \right|.
\end{equation}

\begin{equation}
\mathrm{Precision} = \frac{TP}{\left| M_{\text{pred}} \right|}
\end{equation}

\begin{equation}
    \mathrm{Recall} = \frac{TP}{\left| M_{\text{gold}} \right|}
\end{equation}
\begin{equation}
\mathrm{F1}
=
\frac{2 \cdot \mathrm{Precision} \cdot \mathrm{Recall}}
{\mathrm{Precision}+\mathrm{Recall}}.
\end{equation}

\paragraph{HitRate@K for Nested Financial Terms.}
For the $i$-th instance, let $G_i$ denote the set of gold nested terms and
$P_i^{(K)}$ the top-$K$ predicted terms:
\begin{equation}
\mathrm{HR@K}(i)
=
\frac{\left| G_i \cap P_i^{(K)} \right|}
{\left| G_i \right|}.
\end{equation}

The final score averages over all valid instances:
\begin{equation}
\mathrm{HR@K}
=
\frac{1}{N}
\sum_{i=1}^{N}
\mathrm{HR@K}(i).
\end{equation}

We report HR@1, HR@5, and HR@10 to account for different levels of ranking
tolerance.

\end{document}